\documentclass[10pt,twocolumn,letterpaper]{article}

\usepackage{cvpr}              % To produce the CAMERA-READY version
\usepackage{graphicx}
\usepackage{amsmath}
\usepackage{amssymb}
\usepackage{booktabs}

\usepackage{multirow}
\usepackage{makecell}
\usepackage{bbding}
\usepackage[pagebackref,breaklinks,colorlinks]{hyperref}

\usepackage{algorithm}
\usepackage{algorithmicx}
\usepackage{algpseudocode}
\usepackage[capitalize]{cleveref}
\crefname{section}{Sec.}{Secs.}
\Crefname{section}{Section}{Sections}
\Crefname{table}{Table}{Tables}
\crefname{table}{Tab.}{Tabs.}

\def\cvprPaperID{2377} % *** Enter the CVPR Paper ID here
\def\confName{CVPR}
\def\confYear{2023}

\begin{document}

\title{PSVT: End-to-End Multi-person 3D Pose and Shape Estimation with Progressive Video Transformers}

\author{Zhongwei Qiu$^{1,3,4}$,
        Qiansheng Yang$^2$,
        Jian Wang$^2$,
        Haocheng Feng$^2$,
        Junyu Han$^2$,\\
        Errui Ding$^2$,
        Chang Xu$^3$,
        Dongmei Fu$^{1,4}$,
        Jingdong Wang$^2$
\\
$^1$School of Automation and Electrical Engineering, University of Science and Technology Beijing\\
$^2$Baidu, $^3$University of Sydney, $^4$Beijing Engineering Research Center of Industrial Spectrum Imaging
% \\
% {\tt\small qiuzhongwei@xs.ustb.edu.cn, }
}
% \and

\maketitle

%%%%%%%%% ABSTRACT
\begin{abstract}
Existing methods of multi-person video 3D human Pose and Shape Estimation (PSE) typically adopt a two-stage strategy, which first detects human instances in each frame and then performs single-person PSE with temporal model. However, the global spatio-temporal context among spatial instances can not be captured. In this paper, we propose a new end-to-end multi-person 3D \textbf{P}ose and \textbf{S}hape estimation framework with progressive \textbf{V}ideo \textbf{T}ransformer, termed PSVT. In PSVT, a spatio-temporal encoder (STE) captures the global feature dependencies among spatial objects. Then, spatio-temporal pose decoder (STPD) and shape decoder (STSD) capture the global dependencies between pose queries and feature tokens, shape queries and feature tokens, respectively. 
To handle the variances of objects as time proceeds, a novel scheme of progressive decoding is used to update pose and shape queries at each frame. 
Besides, we propose a novel pose-guided attention (PGA) for shape decoder to better predict shape parameters. 
The two components strengthen the decoder of PSVT to improve performance. Extensive experiments on the four datasets show that PSVT achieves stage-of-the-art results.
\end{abstract}

%%%%%%%%% BODY TEXT
\section{Introduction}
\label{sec:intro}

Multi-person 3D human Pose and Shape Estimation (PSE) from monocular video requires localizing the 3D joint coordinates of all persons and reconstructing their human meshes (e.g. SMPL~\cite{loper2015smpl} model).
As an essential task in computer vision, it has many applications including human-robot interaction detection~\cite{li2020detailed}, virtual reality~\cite{parger2021unoc}, and human behavior understanding~\cite{gatt2019detecting}, etc. Although remarkable progress has been achieved in PSE from videos~\cite{choi2021beyond,yuan2022glamr,wei2022capturing,qiu2022ivt} or images~\cite{sun2021monocular,sun2022putting,choi2022learning}, capturing multi-person spatio-temporal relations of pose and shape simultaneously is still challenging since the difficulty in modeling long-range global interactions.

\begin{figure}[t]
  \centering
%   \fbox{\rule{0pt}{2in} \rule{0.95\columnwidth}{0pt}}
  \includegraphics[width=\linewidth]{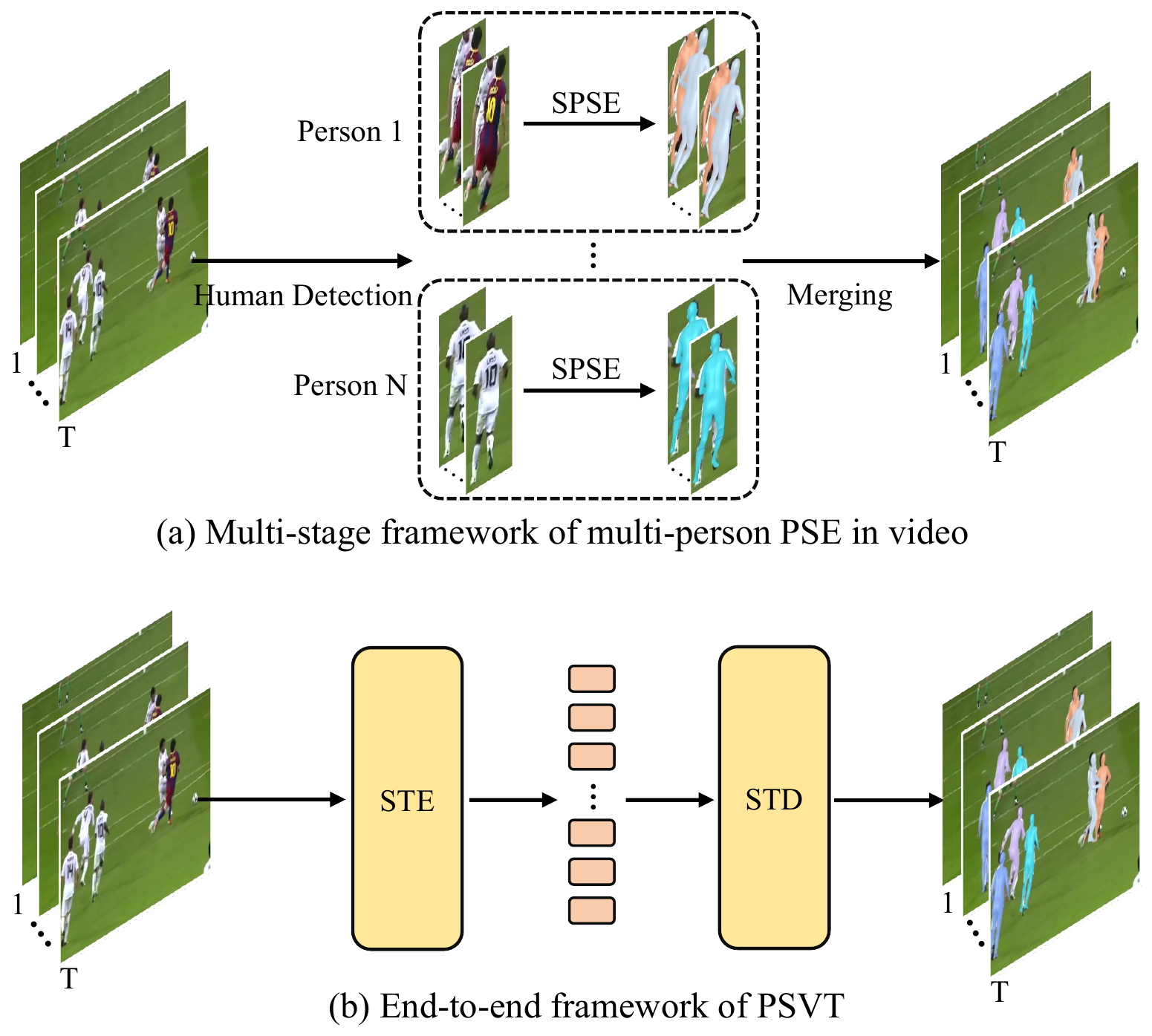}

  \caption{Comparison of multi-stage and end-to-end framework. (a) Existing video-based methods~\cite{kocabas2020vibe,choi2021beyond,wan2021encoder,wei2022capturing} perform single-person pose and shape estimation (SPSE) on the cropped areas by temporal modeling, such as Gated Recurrent Units (GRUs). (b) PSVT achieves end-to-end multi-person pose and shape estimation in video with spatial-temporal encoder (STE) and decoder (STD).}
  \label{fig:teaser}
\end{figure}

To tackle this challenge, as shown in Figure \ref{fig:teaser} (a), existing methods~\cite{kocabas2020vibe,choi2021beyond,yuan2022glamr,wei2022capturing} employ a detection-based strategy of firstly detecting each human instance, then cropping the instance area in each frame and feeding it into the temporal model, such as the recurrent neural network~\cite{doersch2019sim2real,kocabas2020vibe,choi2021beyond}. However, this framework can not capture the spatial relationship among human instances in an image and has the limitation of extracting long-range global context. Besides, the computational cost is expensive since it is proportional to the number of instances in image and it needs extra tracker~\cite{wei2022capturing} to identify each instance. Other temporal smoothing methods~\cite{veges2020temporal,kanazawa2019learning} adopt a post-processing module to align the shape estimated by image-based PSE approaches~\cite{kanazawa2019learning,sun2021monocular,choi2022learning,sun2022putting,li2022cliff}. However, they can not capture temporal information directly from visual image features and lack the ability of long-range global interactions. These multi-stage methods split space and time dimensions and can not be end-to-end optimized.

To strengthen the long-range modeling ability, recently developed Transformer models~\cite{vaswani2017attention,dosovitskiy2020vit} have been introduced in PSE. The Transformer-based mesh reconstruction approaches~\cite {lin2021end,lin2021mesh,zanfir2021thundr,pavlakos2022human} take each human joint as a token and capture the relationship of human joints by attention mechanism. However, the global context among different persons in spatio-temporal dimensions has not been explored. 
Other Transformer-based human pose estimation approaches~\cite{liu2020attention,zheng20213d} explore the spatio-temporal context of human joints for single-person, but not on the  multi-person mesh. Besides, these methods focus on capturing the relations among human joints, while ignoring the relations between human poses and shapes.

% Thus, a unified framework of capturing long-range spatio-temporal context for multi-person 3D human pose and shape estimation in video needs to be established.

To tackle the above problems, we propose an end-to-end multi-person 3D \textbf{P}ose and \textbf{S}hape estimation framework with \textbf{V}ideo \textbf{T}ransformer, termed PSVT, to capture long-range spatio-temporal global interactions in the video.
As shown in Figure \ref{fig:teaser} (b), PSVT formulates the human instance localization and fine-grained pose and mesh estimation as a set prediction problem as \cite{carion2020end,shi2022end}. 
First, PSVT extracts a set of spatio-temporal tokens from the deep visual features and applies a spatio-temporal encoder (STE) on these visual tokens to learn the relations of feature tokens.
Second, given a set of pose queries, a progressive spatio-temporal pose decoder (STPD) learns to capture the relations of human joints in both spatial and temporal dimensions.  
Third, with the guidance of pose tokens from STPD, a progressive spatio-temporal shape decoder (STSD) learns to reason the relations of human mesh and pose in both spatial and temporal dimensions and further estimates the sequence 3D human mesh based on the spatio-temporal global context. Compared with previous shape estimation works~\cite{choi2021beyond,yuan2022glamr,wei2022capturing,sun2021monocular,sun2022putting,choi2022learning}, PSVT achieves end-to-end multi-person 3D pose and shape estimation in video.

In PSVT, different from previous methods, we propose a novel progressive decoding mechanism (PDM) for sequence decoding and pose-guided attention (PGA) for decoder. PDM takes the output tokens from the last frame as the initialized queries for next frame, which enables better sequence decoding for STPD and STSD. PGA aligns the pose tokens and shape queries and further computes the cross-attention with feature tokens from encoder. With the guidance of pose tokens, shape estimation can be more accurate. Our contributions can be summarized as follows:
\begin{itemize}
    \item We propose a novel video Transformer framework, termed PSVT, which is the first end-to-end multi-person 3D human pose and shape estimation framework with video Transformer.
    \item We propose a novel progressive decoding mechanism (PDM) for the decoder of video Transformer, which updates the queries at each frame in the attention block to improve the pose and shape decoding.
    \item We propose a novel pose-guided attention (PGA), which can capture the spatio-temporal relations among pose tokens, shape tokens, and feature tokens to improve the performance of shape estimation.
    \item Extensive experiments on four benchmarks show that PSVT achieves new state-of-the-art results.
\end{itemize}

\section{Related Work}
\label{sec:related_work}
\subsection{Imaged-based 3D Human PSE}
Image-based 3D human PSE methods~\cite{kanazawa2018end,qiu2019learning,zhang2021pymaf,wan2021encoder,choi2022learning,sun2021monocular,qiu2023weakly} estimate the 3D pose, shape, and camera parameters from single RGB image, further to decode human mesh by SMPL model~\cite{loper2015smpl}. They can be divided into single-person and multi-person methods.

\textbf{Single-Person.}
HMR~\cite{kanazawa2018end} proposes an end-to-end single-person human mesh recovery framework by estimating the parameters of the SMPL model. Based on SMPL model, 2D heatmaps and silhouettes are used as the prior information to improve mesh estimation by ~\cite{pavlakos2018learning}. Following this framework, more prior knowledge~\cite{zhang2021pymaf} or stronger backbone network~\cite{wan2021encoder} are adopted to improve the performance of shape estimation. To deal with the occlusion problem, 3DCrowdNet~\cite{choi2022learning} estimates robust 3D human mesh from in-the-wild crowded scenes by using 2D keypoints heatmaps as the key cues.
Although these methods achieve great performance on 3D human PSE, the performances of these methods rely on the accuracy of human detection.

\textbf{Multi-Person.}
Most existing multi-person methods~\cite{kolotouros2019learning,pavlakos2019texturepose,zhang2021pymaf,choi2022learning,qiu2022dynamic} adopt a multi-stage framework to tackle multi-person problem, which firstly conducts human detection and follows a single-person mesh estimation model. However, the multi-stage framework is low-efficiency and can not be optimized in an end-to-end fashion. To solve this problem, ROMP~\cite{sun2021monocular} achieves a novel one-stage pipeline of multi-person image 3D human pose and shape estimation, which directly estimates multiple 2D maps for 2D human detection, positioning, and mesh parameter regression. To capture the relative depth of multiple persons in image, BEV~\cite{sun2022putting} uses an additional
imaginary Bird’s-Eye-View representation to explicitly reason about depth. The single-stage methods~\cite{sun2021monocular,sun2022putting} are high-efficiency, but cannot handle the small object well. 

Although the image-based 3D human pose and shape estimation methods achieve remarkable progress, they lack competitiveness compared with video-based methods because temporal information is important to improve the performance of 3D pose and shape estimation.

\subsection{Video-based 3D Human PSE}
Video-based 3D human pose and shape estimation methods~\cite{kocabas2020vibe,choi2021beyond,wei2022capturing,yuan2022glamr,wan2021encoder} can extract more temporal context to keep the consistency of pose and shape on the time dimension. 
Usually, these video-based methods follow a pipeline of first detecting human instances, and then conducting temporal modeling for a single instance by temporal models, such as recurrent neural networks and 3D convolution.
Typically, VIBE~\cite{kocabas2020vibe} builds a temporal encoder by bidirectional gated recurrent units (GRU) to encode the static feature from the backbone network into temporal features, further to regress SMPL parameters by a regressor. TCMR~\cite{choi2021beyond} uses a GRU-based temporal encoder to extract temporal information with three different encoding strategies. GLAMR~\cite{yuan2022glamr} formulates a four-stage framework with global motion optimization to tackle the occlusion problem in video human mesh recovery. Besides,
MPS-Net~\cite{wei2022capturing} captures human motion among different temporal frames and fuses these motion features to estimate pose and mesh parameters by attention-based feature extraction and integration modules.

These video-based methods are multi-stage and perform single-person temporal modeling in the pose and shape estimation stage. 
Despite the improvements that have been achieved by these temporal models, they are limited to the human detector and cannot be optimized by an end-to-end scheme. Moreover, they cannot capture the depth information between different human instances due to crop images.

\subsection{Transformers in 3D Human PSE}
Recently, Transformer-based models~\cite{lin2021end,lin2021mesh,wan2021encoder,yuan2022glamr,kocabas2021pare,shi2022end} have been introduced in human pose and shape estimation since their strong long-term modeling capabilities of sequence. They can be divided into two categories: capturing joint relations~\cite{lin2021end,lin2021mesh,kocabas2021pare} and capturing temporal relations~\cite{wan2021encoder,yuan2022glamr}. METRO~\cite{lin2021end} and Mesh Graphformer~\cite{lin2021mesh} design end-to-end single-person Transformer and Graph-based Transformer to capture the human joints relations, respectively. However, they cannot process the video problem. MEAD~\cite{wan2021encoder} uses a spatio-temporal Transformer to handle the spatio-temporal relations of human joints. Besides, GLAMR~\cite{yuan2022glamr} uses the attention mechanism to deal with the occlusion problem in video. 
However, existing Transformer-based methods only focus on the single-person pose and estimation problem, and use attention to capture the relations of human joints in spatial or temporal dimensions. Moreover, they cannot capture the relation of multiple human instances in video and their multi-stage framework is expensive. In this paper, we propose a novel end-to-end Transformer-based multi-person human pose and shape estimation framework to capture the relation of human instances in global spatial and temporal dimensions. Besides, we propose new pose-guided attention to decode the shape parameters of human instances.

\begin{figure*}[t]
  \centering

  \includegraphics[width=\linewidth]{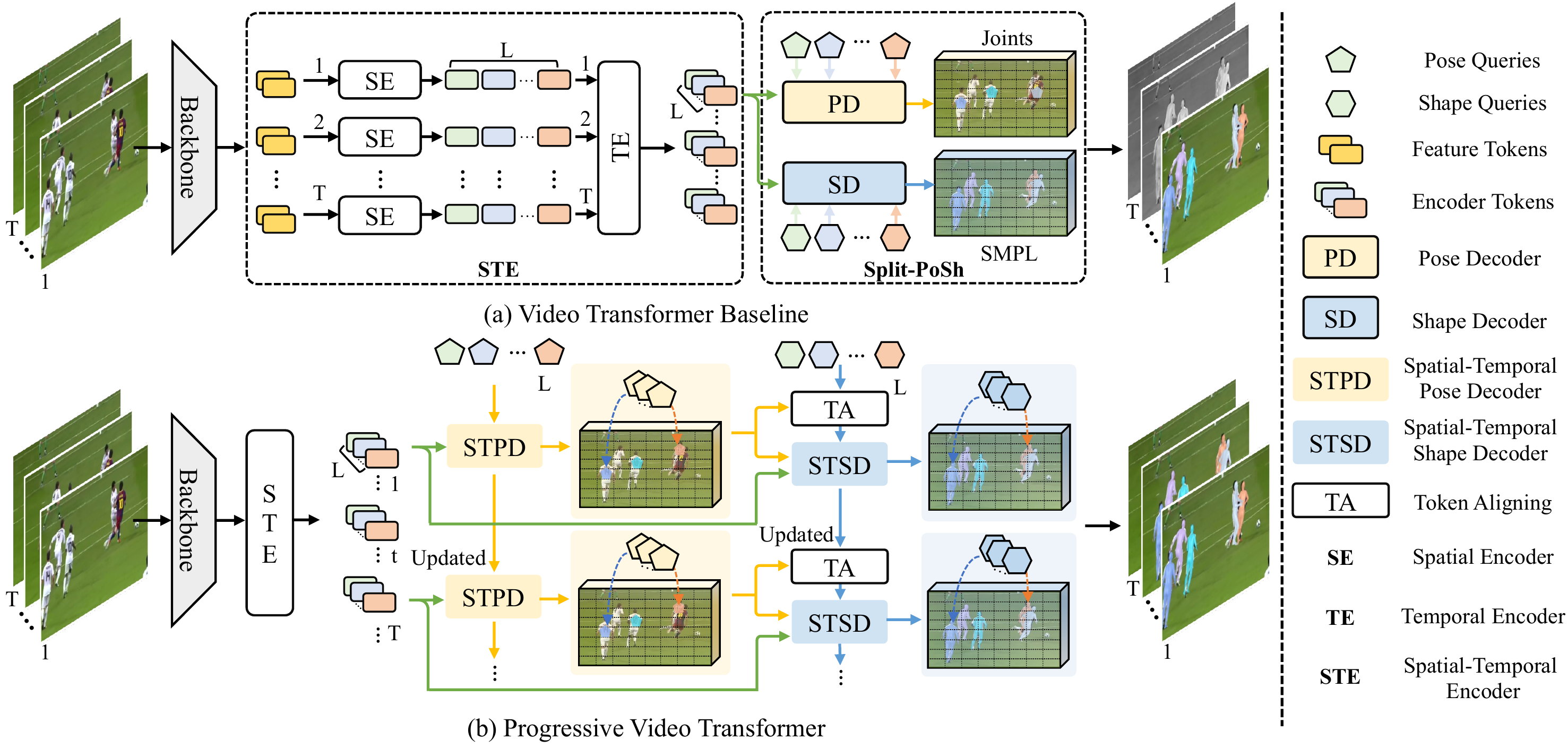}

  \caption{The overview of (a) Video Transformer Baseline (VTL) and (b) Progressive Video Transformer for Pose and Shape estimation (PSVT). 
  A spatial-temporal encoder captures global feature interactions for the decoders of VTL and PSVT. 
  VTL splits the pose decoder and shape decoder (Split-PoSh) to localize human joints and regress the shape parameter of the SMPL model.
  PSVT adopts a progressive decoding mechanism and pose-guided shape decoder.
  }
  \label{fig:framework}
\end{figure*}

\section{Method}
\label{sec:method}
In this section, we first introduce 3D human shape reconstruction models SMPL~\cite{loper2015smpl} and SMPL+A~\cite{patel2021agora}. Then, we propose PSVT, 3D human \textbf{P}ose and \textbf{S}hape estimation with progressive \textbf{V}ideo \textbf{T}ransformer. 
We build a video transformer baseline with vanilla attention. After that, we extend the baseline to a progressive video transformer with progressive decoding mechanism and pose-guided shape attention.

\subsection{SMPL-based Pose and Shape Estimation}
\label{sec:smpl}
SMPL~\cite{loper2015smpl} is a widely-used 3D body shape estimation method, which parameterizes human mesh into low-dimensional parameters. SMPL has been extended to SMIL~\cite{hesse2018learning} and SMPL+A~\cite{patel2021agora} to tackle infant and age problems. Following \cite{patel2021agora}, we adopt the SMPL+A model, which can output human 3D mesh $\mathcal{M}(\theta, \beta, \alpha) \in \mathbb{R}^{6890\times 3}$ with 3D pose $\theta$, shape $\beta$, and age offset $\alpha \in [0,1]$. The pose parameters $\theta \in \mathbb{R}^{6\times 22}$ include the 6D rotations of the 22 body joints in the SMPL+A model. The parameters $\beta \in \mathbb{R}^{10}$ are the top-10 PCA coefficients of shape space. 

Given the a video $V=\{I^t|t\in [1,T]\}$ including $T$ frames, where $I^t \in \mathbb{R}^{H\times W \times 3}$ means $t^{th}$ frame of height $H$ and width $W$, multi-person video 3D human pose and shape estimation aims to output 3D joints $J=\{J^{t}_{i}|t\in[1,T],i\in [1,N]\}$ and mesh $\mathcal{M}=\{\mathcal{M}^t_i|t\in[1,T],i\in [1,N]\}$ for each instance in the video, where $N$ represents the number of instances in an image. $J^{t}_i \in \mathbb{R}^{K\times 3}$ represents the 3D joints of person $i$ in $t^{th}$ frame, which can be generated as $J^{t}_i = \mathcal{W}\mathcal{M}^t_i$, where $\mathcal{W} \in \mathbb{R}^{K\times 6890}$ is weight matrix to map mesh into joints. $K$ is the number of joints. 

\subsection{Video Transformer Baseline}
\label{sec:vtb}

Different from existing multi-stage video-based methods~\cite{kocabas2020vibe,choi2021beyond,wan2021encoder,wei2022capturing,yuan2022glamr}, we propose an end-to-end multi-person 3D pose and shape estimation framework, which consists of backbone network, spatio-temporal encoder, and decoder.

\textbf{Framework.} 
The framework of the video transformer baseline is shown in Figure \ref{fig:framework} (a). 
Given the input video $V$, deep features $F^t \in \mathbb{R}^{\frac{H}{s} \times \frac{W}{s} \times C}$ are extracted for each frame by backbone network HRNet~\cite{sun2019deep}, where $s$ and $C$ represent scale and feature dim, respectively. 
Then, $F^t$ are extracted as tokens $\tau^t_e \in \mathbb{R}^{L\times D}$ by patch embedding with a patch size of $B \times B$. $L$ represents the number of tokens and $D$ is embedding dim.
For the video including $T$ frames, $T \times L$ tokens $\tau_e=\{\tau^t_e|t\in [1,T]\}$ are generated. 
These tokens are sent into a spatio-temporal encoder. After that, $L$ pose queries and $L$ shape queries for each frame, which have the same spatial positions with tokens $\tau^t_e$, are sent into the pose decoder and shape decoder to reason joints heatmaps and mesh parameters, respectively.

\textbf{Encoder.}
The feature maps $F$ are extracted as tokens $\tau_e \in \mathbb{R}^{T\times L \times D}$, which serve as the input of the transformer encoder. To capture the relations among tokens in both spatial and temporal dimensions and reduce the computational costs, we adopt divided attention for video encoder, which firstly computes the self-attention among the $L$ spatial tokens and then follows the self-attention among $T$ tokens in the same spatial place. This scheme is termed Spatio-Temporal Encoder (\textbf{STE}) since it captures the global context in both spatial and temporal dimensions.

For the self-attention of spatio-temporal encoder, we adopt the vanilla multi-head attention following \cite{vaswani2017attention}. Let $A$ denote basic attention, and it is computed as:
\begin{equation}
\label{eq:att}
    A(\mathcal{Q}, \mathcal{K}, \mathcal{V}) = softmax(\frac{\mathcal{Q}\cdot \mathcal{K}^\top}{\sqrt{d}}) \cdot \mathcal{V},
\end{equation}
where $\mathcal{Q}$, $\mathcal{K}$, $\mathcal{V}$, and $d$ are queries, keys, values, and feature dim of tokens, respectively. Dividing tokens into $h$ groups, then the multi-head attention(MHA) can be formulated as:
\begin{equation}
\label{eq:mha}
\begin{aligned}
\text{MHA}(\mathcal{Q}, \mathcal{K}, \mathcal{V}) & = P(Concat(head_1, ..., head_h)) \\
s.t.~head_i & = A(\mathcal{Q}_i,\mathcal{K}_i,\mathcal{V}_i ), i \in [1,h],
\end{aligned}
\end{equation}
where $P(\cdot)$ represents linear projection function. $Concat$ is the operation of concatenating features along feature axis.

\textbf{Decoder.}
For pose and shape estimation, we build pose decoder and shape decoder with attention as Equation \ref{eq:mha}, termed \textbf{Split-PoSh} since the pose decoder and shape decoder are split.
For a frame $I^t$ in the video, given $L$ pose queries $\mathcal{Q}^t_{pose}$, pose decoder computes the cross-attention between pose queries and feature tokens $\tau^t_e$ from encoder. The attention in pose decoder can be denoted as $\text{MHA}_{pose}(\mathcal{Q}^t_{pose}, \tau^t_e, \tau^t_e)$. The output tokens from attention $\text{MHA}_{pose}$ are used to regress joints 2D heatmaps $M_{2D} \in \mathbb{R}^{K \times \frac{H}{s} \times \frac{W}{s}}$, joints offsets $M_o \in \mathbb{R}^{3\times K \times \frac{H}{s} \times \frac{W}{s}}$, and camera depth map $M_d \in \mathbb{R}^{1\times \frac{H}{s} \times \frac{W}{s}}$ by a multi-layer perceptron. Finally, the tokens at the positions with top $N$ heat scores are used to predict the locations of $N$ persons.

Similar to pose decoder, shape decoder computes the attention $\text{MHA}_{shape}(\mathcal{Q}^t_{shape}, \tau^t_e, \tau^t_e)$ between shape queries and feature tokens, then output shape parameters maps $M_s \in \mathbb{R}^{143\times \frac{H}{s} \times \frac{W}{s}}$. Each token in $M_s$ contains pose parameters of 6D rotations $\theta$, shape parameters $\beta$, and age offset $\alpha$, which is further used to generate SMPL+A mesh. The tokens at the same top $N$ positions are used to regress $\mathcal{M}^t = \{\mathcal{M}^t_i(\theta, \beta, \alpha) | i\in [1,N]\}$ with the help of their corresponding position and depth $(x_i, y_i, d_i)$.
Then, the 3D joints $J^t = \{J^{t}_i|i\in [1,N]\}$ can be obtained as the process introduced in Section \ref{sec:smpl}.

\subsection{Progressive Video Transformer}
\label{sec:psvt}
Compared with video Transformer baseline, progressive video Transformer has two differences. 
1) The sequence decoding process is based on recurrent structure, in which the pose and shape queries in each frame are updated with the prior output tokens from the last step. 
This mechanism enables decoder to capture the global relationship and reduce the computation costs of spatio-temporal attention. Meanwhile, the updated queries can improve the performance of the decoder since the appearance of human instances may change significantly as time proceeds.
2) We propose pose-guided shape attention (PGA) in shape decoder. PGA improves the performance of shape decoder since the depth and joint information in pose tokens can guide shape decoder to generate better mesh parameters. The framework of progressive video Transformer is shown in Figure \ref{fig:framework} (b), which includes backbone network, encoder, and decoder.

\textbf{Progressive Decoding Mechanism.}
In each frame of the video, human instances have different appearances and pose. Therefore, using the same queries to decode pose and shape tokens is not intuitive. To tackle this problem, we adopt the recurrent structure with an attention-based decoder to decode the sequence of pose and shape tokens. 

As shown in Figure \ref{fig:framework} (b), given the initialized pose queries $\mathcal{Q}_{pose}^t$, and features tokens $\tau^t_e$ from STE at $t^{th}$ frame, the pose tokens $\tau^t_{pose}$ are firstly decoded from spatio-temporal pose decoder (STPD) by computing the pose cross-attention between  $\mathcal{Q}_{pose}^t$ and $\tau^t_e$, denoted as 
\begin{equation}
\label{eq:pdm_pose}
\begin{aligned}
    \tau^t_{pose} &= \text{STPD}(\hat{\mathcal{Q}}_{pose}^t, \tau^t_e),\\
    s.t. ~\hat{\mathcal{Q}}_{pose}^t & = \psi(\mathcal{Q}_{pose}^t, \tau^{t-1}_{pose}),
\end{aligned}
\end{equation}
where $\hat{\mathcal{Q}}_{pose}^t$ represents the updated pose queries by fusing the decoded pose tokens $\tau^{t-1}_{pose}$ at $(t-1)^{th}$ frame and the initialized pose queries $\mathcal{Q}_{pose}^t$, $t\in [1,T]$. $\psi(\cdot)$ is the linear projection layer to fuse tokens. $\text{STPD}(\cdot)$ represents the operation of spatio-temporal pose decoder.

For shape decoder, the inputs include shape queries $\mathcal{Q}_{shape}^t$, decoded pose tokens $\tau^t_{pose}$, and feature tokens $\tau^t_e$ from STE. $\mathcal{Q}_{shape}^t$ and $\tau^t_{pose}$ are firstly aligned and then sent into recurrent STSD to decode shape tokens $\tau^t_{shape}$. This process can be formulated as 
\begin{equation}
\label{eq:pdm_shape}
\begin{aligned}
    \tau^t_{shape} & = \text{STSD}(\hat{\mathcal{Q}}^t_{shape}, \tau^t_e),\\
    s.t. ~\hat{\mathcal{Q}}^t_{shape} & = \text{TA}(\psi(\mathcal{Q}^t_{shape}, \tau^{t-1}_{shape}), \tau^t_{pose}),
\end{aligned}
\end{equation}
where $\hat{\mathcal{Q}}^t_{shape}$ represents the updated shape queries, which is computed by token aligning on shape tokens $\psi(\mathcal{Q}^t_{shape}, \tau^{t-1}_{shape})$ and pose tokens $\tau^t_{pose}$. $\text{TA} (\cdot)$ represents the operation of token aligning (TA), which is a cross-attention with shape tokens as queries and pose tokens as keys and values. $\psi(\cdot)$ is the linear projection layer to fuse shape queries and shape tokens at $(t-1)^{th}$ frame,  $t \in [1,T]$. $\text{STSD}(\cdot)$ represents spatio-temporal shape decoder.

\begin{figure}[t]
  \centering
%   \fbox{\rule{0pt}{2in} \rule{0.95\columnwidth}{0pt}}
  \includegraphics[width=\linewidth]{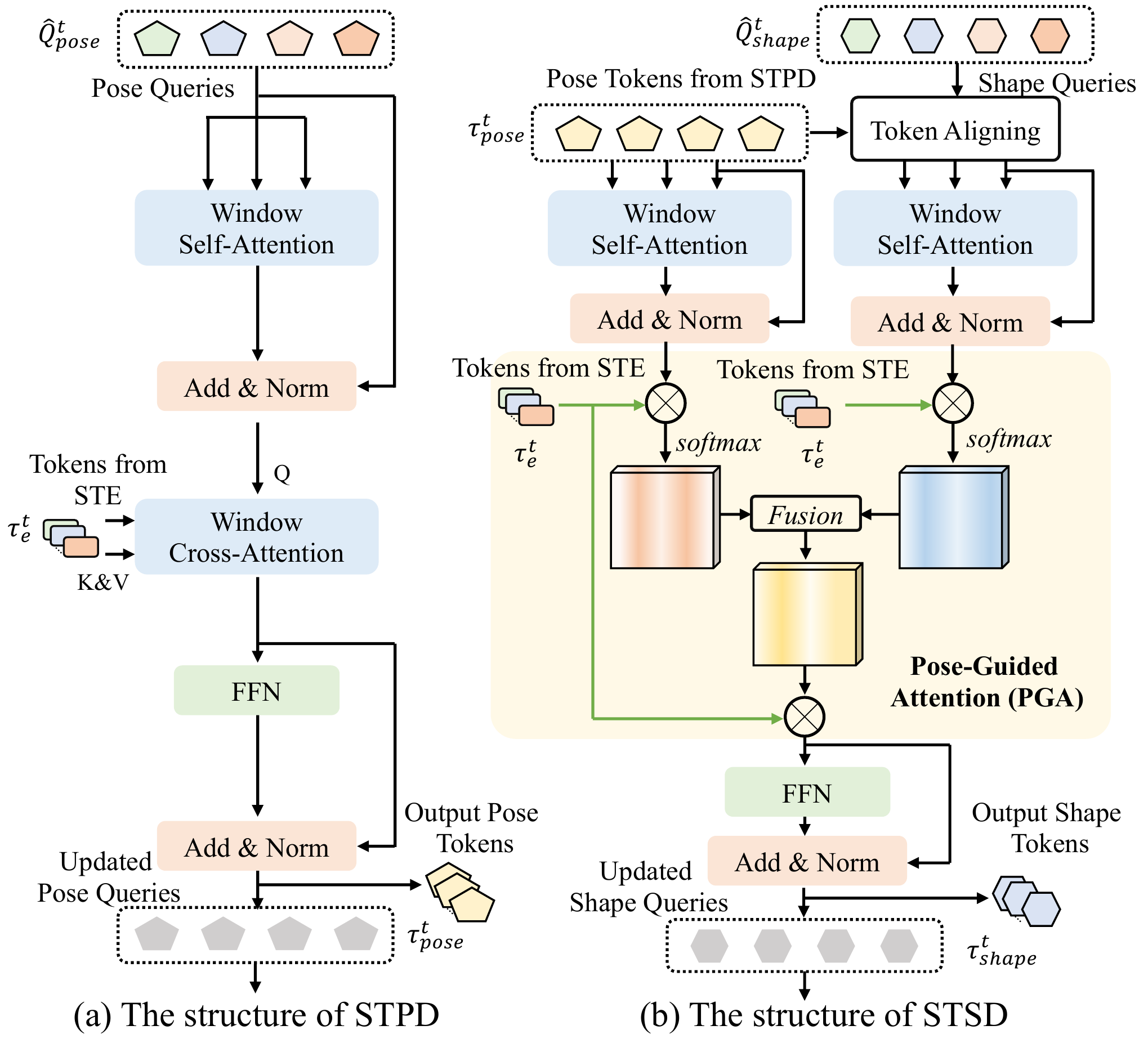}

  \caption{The structure of (a) STPD and (b) STSD. STPD adopts window self-attention~\cite{liu2021swin} to capture the relations between spatial objects and window cross-attention to capture the interactions between pose queries and feature tokens. STPD uses PGA to capture the relations between shape queries and feature tokens with the guidance of pose tokens. $\otimes$ represents matrix multiplication.}
  \label{fig:stpd}
\end{figure}

\textbf{Spatio-Temporal Pose Decoder.}
Given pose queries $\mathcal{Q}_{pose}$ and feature tokens $\tau_{e}$ from STE, Spatio-Temporal Pose Decoder (STPD) aims to decode the pose tokens $\tau_{pose}=\{\tau^t_{pose}|t\in [1,T]\}$. 
The structure of STPD is shown in Figure \ref{fig:stpd} (a).
First, $L$ pose queries are updated to $\hat{\mathcal{Q}}^t_{pose}$ as Equation \ref{eq:pdm_pose}, further are fed into a window-based pose self-attention module to capture the relations among $N$ human instances, which can provide the spatial depth context among these objects. After the pose self-attention, the output tokens $\Tilde{\mathcal{Q}}^t_{pose}$ at $t^{th}$ frame can be generated as
\begin{equation}
    \Tilde{\mathcal{Q}}^t_{pose} = \text{MHA}(\hat{\mathcal{Q}}^t_{pose}, \hat{\mathcal{Q}}^t_{pose}, \hat{\mathcal{Q}}^t_{pose}) +\hat{\mathcal{Q}}^t_{pose} ,
\end{equation}
where $\text{MHA}(\cdot)$ is multi-head self-attention as Equation~\ref{eq:mha}.

Then, output tokens $\Tilde{\mathcal{Q}}^t_{pose}$ are fed into a cross-attention module as queries, which extract the relation between pose queries and feature tokens $\tau^t_e$ from STE. After cross-attention, a feed-forward network (FFN) is used to regress the output tokens $\tau^t_{pose}$. This process can be formulated as 
\begin{equation}
\begin{aligned}
    \tau^t_{pose} &= \text{FFN}(\upsilon_{pose}) + \upsilon_{pose},\\
    s.t. ~\upsilon_{pose} &= \text{MHA}(\Tilde{\mathcal{Q}}^t_{pose}, \tau^t_e, \tau^t_e),
\end{aligned}
\end{equation}
where we ignore the layer norm for simplification. FFN consists of two linear layers. For each frame in video, the output tokens $\tau^t_{pose} \in \mathbb{R}^{L\times D}$ are used to regress 2D heatmaps $M_{2D}$, 3D offsets $M_o$, and camera depth map $M_d$ as the pose decoder in video Transformer baseline.

\textbf{Spatio-Temporal Shape Decoder.}
Given the shape queries $\mathcal{Q}_{shape}$, pose tokens $\tau_{pose}$, and feature tokens $\tau_{e}$ from STE, Spatio-Temporal Shape Decoder (STSD) aims to decode the shape tokens $\tau_{shape}=\{\tau^t_{shape}|t\in [1,T]\}$. Different from pose decoder, a novel Pose-Guided Attention (\textbf{PGA}) module is proposed to decode shape parameters.

The structure of STSD is shown in Figure \ref{fig:stpd} (b). First, $L$ shape queries are updated to $\hat{\mathcal{Q}}^t_{shape}$ by token aligning with pose tokens $\tau^t_{pose}$. Then, a window-based pose self-attention module and a window-based shape self-attention module are applied on updated shape queries and pose tokens to generate shape tokens $\Tilde{\mathcal{Q}}^t_{shape}$ and pose tokens $\Tilde{\tau}^t_{pose}$, respectively. The process can be formulated as 
\begin{equation}
    \Tilde{\eta} = \text{MHA}(\eta, \eta, \eta) + \eta, \eta \in \{\hat{\mathcal{Q}}^t_{shape}, \tau^t_{pose}\},
\end{equation}
where $\eta$ is wildcard character for pose tokens $\tau^t_{pose}$ and shape tokens $\hat{\mathcal{Q}}^t_{shape}$. After pose and shape self-attention, $\Tilde{\mathcal{Q}}^t_{shape}$ and $\Tilde{\tau}^t_{pose}$ are fed into PGA module to compute cross-attention with feature tokens $\tau^t_e$, respectively. The pose and shape cross-attention matrices ($\chi^t_{pose}$, $\chi^t_{shape}$) are fused by linear projection layer $fc$, further to recover the shape tokens. The process of PGA can be formulated as 
\begin{equation}
\begin{aligned}
 \tau^t_{shape} & = \text{FFN}(\upsilon_{shape}) + \upsilon_{shape},\\
 \upsilon_{shape} & = fc(\chi_{pose}^t \oplus \chi_{shape}^t) \otimes \tau^t_{e},\\
 \chi_{pose}^t & = softmax(\Tilde{\tau}^t_{pose} \otimes (\tau^t_e)^\top), \\
  \chi_{shape}^t & = softmax(\Tilde{\mathcal{Q}}^t_{shape} \otimes (\tau^t_e)^\top),
\end{aligned}
\end{equation}
where $\oplus$ indicates concatenating matrices. $\otimes$ is matrix multiplication. For each frame in video, the output tokens $\tau^t_{shape} \in \mathcal{R}^{L\times D}$ are used to regress mesh parameter maps $M_s$ as the shape decoder in video Transformer baseline.

\subsection{Loss Function}
Each token in pose maps can localize one object. Following \cite{sun2021monocular,sun2022putting}, combined with pose maps ($M_{2d}$, $M_o$, and $M_d$) and shape maps $M_s$, we can generate $N$ SMPL mesh $\mathcal{M}(\theta, \beta, \alpha)$ for $N$ instances. For each person, the 3D joints $J$ can be obtain from $\mathcal{M}(\theta, \beta, \alpha)$. Following previous works~\cite{kanazawa2018end,sun2021monocular,sun2022putting}, we use pose losses and mesh losses to supervise outputs. Pose losses include $L_2$ losses of heatmaps $\mathcal{L}_{2D}$, offsets $\mathcal{L}_o$, depths $\mathcal{L}_d$. mesh losses include $L_2$ losses of pose parameters $\mathcal{L}_\theta$, shape parameters $\mathcal{L}_\beta$ and age parameters $\mathcal{L}_\alpha$. Besides, the $L_2$ loss of projected 3D joints $\mathcal{L}_J$ and Mixture Gaussian prior loss of SMPL model in ~\cite{kolotouros2019learning} are also used. The total loss $\mathcal{L}$ is 
\begin{equation}
    \mathcal{L} = W_{pose}\mathcal{L}_{pose} + W_{mesh}\mathcal{L}_{mesh} + W_{J}\mathcal{L}_{J} + W_{p}\mathcal{L}_p,
\end{equation}
where $\mathcal{L}_{pose} = \mathcal{L}_{2D} + \mathcal{L}_{o} +\mathcal{L}_{d}$ and $\mathcal{L}_{mesh} = \mathcal{L}_{\theta} + \mathcal{L}_{\beta}  +\mathcal{L}_{\alpha}$. $W_{pose}$, $W_{mesh}$, $W_J$, and $W_p$ are the weights for pose losses, mesh losses, joints loss, and prior loss, respectively.

\section{Experiments}
\label{sec:exp}
\subsection{Implementation Details}
Following previous works~\cite{sun2022putting,wei2022capturing}, we use HRNet as the backbone network. For training, we use the two-stage training strategy as \cite{sun2022putting}. For the first stage, PSVT is pre-trained on two basic 3D datasets (Human3.6M~\cite{ionescu2013human3} and Muco-3DHP~\cite{mehta2018single}) and four 2D datasets (COCO~\cite{lin2014microsoft}, MPII~\cite{andriluka20142d}, LSP~\cite{johnson2011learning}, and CrowdPose~\cite{li2019crowdpose}). Then, PSVT is finetuned on the target 3D datasets (RH~\cite{sun2022putting}, AGORA~\cite{patel2021agora}, 3DPW~\cite{von2018recovering}) with loading the pre-trained parameters for better convergence. When training, PSVT is trained with 120 epochs for first-stage training and 60 epochs for second-stage training. Learning rate is 5e-5. Batch size is 64 and the input size of image is $512\times 512$. Following~\cite{sun2021monocular}, loss weights $W_{pose}$, $W_{mesh}$, $W_{J}$, $W_{p}$ are set to 160, 1, 360, and 1.6, respectively. 8 Tesla V100 GPUs are used for training.

\subsection{Datasets and Evaluation Metrics}
\textbf{RH dataset.}
RH dataset is a multi-person dataset \cite{sun2022putting} for evaluating depth reasoning, which includes about 7.6K images with weak annotations of over 24.8K people. 
Following ~\cite{sun2022putting}, the percentage of correct depth relations (PCDR$^{0.2}$) with a threshold of 0.2m is used as the metric to evaluate the accuracy of depth reasoning.

\textbf{AGORA dataset.}
AGORA dataset is a multi-person dataset~\cite{patel2021agora}, which contains 14K training images and 3D testing images with accurate annotations of body meshes and 3D translations. It contains 4240 high-realism textured scans since the images are synthetic.
For evaluation, mean per joint position error (MPJPE) and mean vertex error (MVE) are used to evaluate the accuracy of 3D pose and mesh estimation. Considering the missing detections, normalized mean joints error (NMJE) and normalized mean vertex error (NMVE) are also used for evaluation.

\textbf{CMU Panoptic dataset.}
CMU Panoptic~\cite{joo2015panoptic} is a large-scale multi-person dataset for 3D human pose and shape estimation. The images are captured by multiple cameras. Following previous works~\cite{kocabas2020vibe,sun2021monocular,sun2022putting}, we evaluate PSVT on the testing set of CMU Panoptic without using training set for fair comparison. The evaluation metric is MPJPE.

\textbf{3DPW dataset.}
3DPW~\cite{von2018recovering} is a outdoor multi-person dataset, which contains 22K images and 35K images for training and testing, respectively. For comparison with previous works~\cite{wei2022capturing,sun2022putting,choi2021beyond,kocabas2020vibe}, PA-MPJPE, MPJPE, and MPVE are used as evaluation metrics.

\subsection{Comparison with the State-of-the-art Methods}
We compare PSVT with existing methods on two image-based and video-based benchmarks.

\textbf{Evaluation on RH Dataset.}
To evaluate the monocular depth reasoning ability of PSVT on the in-the-wild images, we compare PSVT with other SOTA methods on RH~\cite{sun2022putting} dataset. The results are shown in Table \ref{tab:rh}. Compared with single-person method~\cite{moon2019camera} and other multi-person methods~\cite{jiang2020coherent,zhen2020smap,sun2021monocular,sun2022putting}, PSVT achieves a PCDR$^{0.2}$ of 71.23\% and outperforms them on each category (Baby, Kid, Teenager, and Adult). Although BEV~\cite{sun2021monocular} uses the 3D representation combined with bird's eye view features to enhance the depth reasoning ability, PSVT has stronger depth reasoning ability than BEV since the pose-guided attention, which captures the global spatial context between different human instances.

\begin{table}
  \centering
  \renewcommand\tabcolsep{10pt}
  \resizebox{\columnwidth}{!}{
  \begin{tabular}{@{}l|ccccc@{}}
    \toprule
    \multirow{2}{*}{Method} & \multicolumn{5}{c}{PCDR$^{0.2}$($\% $)$\uparrow$} \\
    \cmidrule{2-6}
     & Baby & Kid & Teen & Adult & All\\
    \midrule
    3DMPPE~\cite{moon2019camera} & 39.33 & 51.42 & 60.91 & 57.95 & 57.47 \\
    CRMH~\cite{jiang2020coherent} & 34.74 & 48.37 & 59.11 & 55.47 & 54.83 \\
    SMAP~\cite{zhen2020smap} & 31.58 & 40.29 & 47.35 & 41.65 & 41.55 \\
    ROMP~\cite{sun2021monocular} & 30.08 & 48.41 & 51.12 & 55.34 & 54.81 \\
    BEV~\cite{sun2022putting}  & 60.77 & 67.09 & 66.07 & 69.71 & 68.27 \\
    \midrule

    \textbf{PSVT} (Ours) & \textbf{64.00} & \textbf{71.29} & \textbf{70.45} & \textbf{71.95} & \textbf{71.23}\\
    \bottomrule
  \end{tabular}
  }
  \caption{The comparison of accuracy on RH~\cite{sun2022putting} dataset.}
  \label{tab:rh}
\end{table}

\textbf{Evaluation on AGORA Dataset.}
To evaluate the 3D pose and mesh estimation, we test PSVT on the AGORA dataset as previous works~\cite{kocabas2021pare,patel2021agora,kocabas2021spec,sun2021monocular,sun2022putting}.
As shown in Table \ref{tab:agora}, for the matched objects, PSVT achieves 94.1 and 97.7 in MVE and MPJPE, respectively. For all objects, PSVT achieves 101.2 and 105.1 in NMVE and NMJPE, respectively.
Compared with SOTA method~\cite{sun2022putting}, PSVT outperforms BEV on all objects by 6.6\% and 7.2\% in NMVE and NMJE, respectively. 
These results demonstrate the effectiveness of PSVT on pose and mesh estimation.

\begin{table}
  \centering
  \renewcommand\tabcolsep{14pt}
  \resizebox{\columnwidth}{!}{
  \begin{tabular}{@{}l|cc|cc@{}}
    \toprule
    \multirow{2}{*}{Method} & \multicolumn{2}{c|}{Matched$\downarrow$} & \multicolumn{2}{c}{All$\downarrow$}\\
    \cmidrule{2-5}
     & MVE & MPJPE & NMVE & NMJE \\
    \midrule
    PARE~\cite{kocabas2021pare} & 140.9 & 146.2 & 167.7 & 174.0 \\
    SPIN~\cite{patel2021agora} & 148.9 & 153.4 & 193.4 & 199.2 \\
    SPEC~\cite{kocabas2021spec} & 106.5 & 112.3 & 126.8 & 133.7 \\
    ROMP~\cite{sun2021monocular} & 103.4 & 108.1 & 113.6 & 118.8 \\
    BEV~\cite{sun2022putting}  & 100.7 & 105.3 & 108.3 & 113.2\\
    \midrule
    \textbf{PSVT} (Ours) & \textbf{94.1} & \textbf{97.7} & \textbf{101.2} & \textbf{105.1} \\
    \bottomrule
  \end{tabular}
  }
  \caption{The comparison of mean errors on AGORA~\cite{patel2021agora} dataset.}
  \label{tab:agora}
\end{table}

\textbf{Evaluation on CMU Dataset.}
To evaluate the effectiveness of PSVT on videos, we compare PSVT with other SOTA methods on CMU dataset~\cite{joo2015panoptic}. As shown in Table~\ref{tab:cmu}, PSVT outperforms previous works and achieves 105.7mm in MPJPE, which has relative gain of 3.5\% and shows the stronger generalization ability of PSVT since PSVT is not been trained on the CMU dataset.

\begin{table}
  \centering
  \renewcommand\tabcolsep{8pt}
  \resizebox{\columnwidth}{!}{
  \begin{tabular}{@{}l|cccc|c@{}}
    \toprule
    Method & Haggl. & Mafia & Ultim. & Pizza & Mean$\downarrow$\\
    \midrule
    MubyNet~\cite{zanfir2018deep} & 141.4 & 152.3 & 145.0 & 162.5 & 150.3 \\
    MSC~\cite{zanfir2018monocular}& 140.0 & 165.9 & 150.7 & 156.0 & 153.4 \\
    CRMH~\cite{jiang2020coherent} & 129.6 & 133.5 & 153.0 & 156.7 & 143.2 \\
    ROMP~\cite{sun2021monocular}  & 110.8 & 122.8 & 141.6 & 137.6 & 128.2 \\
    3DCrowdNet~\cite{choi2022learning}& 109.6 & 135.9 & 129.8 & 135.6 & 127.3\\
    BEV~\cite{sun2022putting}  & 90.7 & 103.7 & \textbf{113.1} & 125.2 & 109.5 \\
    \midrule

    \textbf{PSVT} (Ours) & \textbf{88.7} & \textbf{97.9} & 115.2 & \textbf{121.1} & \textbf{105.7}\\
    \bottomrule
  \end{tabular}
  }
  \caption{The comparison of MPJPE on CMU~\cite{joo2015panoptic} dataset.}
  \label{tab:cmu}
\end{table}

\textbf{Evaluation on 3DPW Dataset.}
We compare PSVT with other methods on the 3DPW dataset~\cite{von2018recovering}. Without using temporal information, PSVT outperforms all image-based methods and achieves 45.7, 75.5, and 84.9 in PA-MPJPE, MPJPE, and MPVE, respectively. Moreover, PSVT achieves new SOTA results by using 9 frames. Compared with Transformer-based methods~\cite{kocabas2021pare,wei2022capturing,yuan2022glamr,wan2021encoder}, PSVT achieves better results in shape estimation (84.0 in MPVE) since PGA and progressive decoding mechanism.

\begin{table}
  \centering
  \renewcommand\tabcolsep{8pt}
  \resizebox{\columnwidth}{!}{
  \begin{tabular}{@{}l|c|c|ccc@{}}
    \toprule
    Method & Frame & Type & PA-MPJPE & MPJPE & MPVE \\
    \midrule
    PyMAF~\cite{zhang2021pymaf} &\multirow{6}{*}{1}  & \multirow{4}{*}{SP} &58.9 & 92.8 & 110.1\\
    HybrIK~\cite{li2021hybrik} &&  & 48.8 & 80.0 & 94.5 \\
    METRO$^*$~\cite{lin2021end} & &  & 47.9 & 77.1 & 88.2 \\
    PARE$^*$~\cite{kocabas2021pare} & &  & 46.5 & \textbf{74.5} & 88.6 \\
    % MGraphormer$^*$~\cite{lin2021mesh} & &  & 45.6 & 74.7 & 87.7 \\
    \cmidrule{3-6}
 
    ROMP~\cite{sun2021monocular} & & \multirow{2}{*}{MP} &47.3 & 76.7 & 93.4 \\
    BEV~\cite{sun2022putting} & &  & 46.9 & 78.5 & 92.3 \\
    \midrule
    \textbf{PSVT} (Ours) & 1 & MP & \textbf{45.7} & 75.5 & \textbf{84.9}\\
    \midrule

    HMMR~\cite{kanazawa2019learning} & 20 & \multirow{8}{*}{SP} & 72.6 & 116.5 & 139.3 \\
    MEVA~\cite{luo20203d} & 90 &  & 54.7 & 86.9 & - \\
    VIBE~\cite{kocabas2020vibe} & 16 &  & 56.9 & 90.2 & 109.5 \\
    TCMR~\cite{choi2021beyond} & 16&  & 52.7 & 86.5 & 102.9 \\
    MPS-Net$^*$~\cite{wei2022capturing} & 16 &  & 52.1 & 84.3 & 99.7\\
    VIBE+$D_m$~\cite{kocabas2020vibe} & 16 &  & 51.9 & 82.9 & 99.1 \\
    GLAMR$^*$~\cite{yuan2022glamr} & 75 &  & 51.1 & - & -\\
    MAED$^*$~\cite{wan2021encoder} & 16 &  &45.7 & 79.1 & 92.6 \\
    \midrule

    \textbf{PSVT} (Ours) & 9 & MP & \textbf{43.5} & \textbf{73.1} & \textbf{84.0}\\
    \bottomrule
  \end{tabular}
  }
  \caption{The results on 3DPW~\cite{von2018recovering} dataset. $*$ indicates Transformer-based method. SP means single-person method. MP means multi-person method, which input is the whole image including multiple persons. MP is more difficult and efficient.}
  \label{tab:3dpw}
\end{table}

\textbf{Params, FLOPs and Model Size.}
The comparisons of parameters, FLOPs, and model size are shown in Table \ref{tab:params}. Compared with the single-person video-based method MPS-Net~\cite{wei2022capturing}, PSVT achieves better performance with a smaller model size and parameter. Besides, PSVT can output multi-person poses and shapes from video in one stage while other video-based methods need two stages. Compared with BEV~\cite{sun2022putting}, PSVT achieves better results with comparable parameters and computational costs.

\begin{table}
  \centering
  \renewcommand\tabcolsep{4pt}
  \resizebox{\columnwidth}{!}{
  \begin{tabular}{@{}l|c|ccc|c@{}}
    \toprule
    Method & Type & Params (M) & FLOPs (G) & Model (MB)& PA-MPJPE  \\
    \midrule
    VIBE~\cite{kocabas2020vibe}& SP & 72.43 & 42.17& 776 & 56.9 \\
    MEVA~\cite{luo20203d} & SP & 85.72 & 42.46& 858.8 & 54.7 \\
    TCMR~\cite{choi2021beyond} & SP& 108.89 & 42.99& 1073 & 52.7 \\
    MPS-Net~\cite{wei2022capturing} & SP& 39.63 & 42.45 & 331 & 52.1 \\
    BEV~\cite{sun2022putting} & MP &35.86 & 48.89 & 144 & 46.9\\
    \midrule
    \textbf{PSVT} (Ours) & MP& 38.36 & 68.67 & 164 & 43.5\\
    \bottomrule
  \end{tabular}
  }
  \caption{Comparison of FLOPs, Parameters, and model size. 
 FLOPs is computed with image size of $512\times 512$ and backbone network of HRNet-32. The FLOPs of SP methods will increase manyfold with the human number in the image.
  }
  \label{tab:params}
\end{table}

\subsection{Ablation Study}

\textbf{Different Encoders and Decoders}
The ablation study of different encoders and decoders is shown in Table \ref{tab:aba}. With the same spatial encoder, PSVT with PGA achieves better results of 45.7 in PA-MPJPE, 75.5 in MPJPE, and 84.9 in MPVE. Compared with Split-PoSh in video Transformer baseline, the PGA in PSVT brings relative improvements of 3.2 \%, 4.9\%, and 4.2\% in PA-MPJPE, MPJPE, and MPVE, respectively.
With spatio-temporal encoder, the progressive decoding mechanism and PGA enable PSVT to learn more temporal information, which helps PSVT achieve 43.5 in PA-MPJPE, 73.1 in MPJPE, and 84.0 in MPVE. These results verify the effectiveness of the proposed PDM and PGA in PSVT.

\begin{table}
  \centering
  \renewcommand\tabcolsep{3pt}
  \resizebox{\columnwidth}{!}{
  \begin{tabular}{@{}c|c|cl|ccc@{}}
    \toprule
    \# & Type& Encoder & \multicolumn{1}{l|}{Decoder} & PA-MPJPE & MPJPE & MPVE \\
    \midrule
    1 & \multirow{4}{*}{Image} &\XSolidBrush & Conv & 49.8 & 86.2 & 96.9\\
    2 &  & S & Conv & 48.9 & 84.1 & 92.5\\
    3 &  & S & Split-PoSh & 47.2 & 79.4 & 88.6\\
    4 &  & S & PGA &45.7 & 75.5 & 84.9\\
    \midrule
    5 & \multirow{4}{*}{Video} & ST & Conv & 48.2 & 83.4 & 90.3\\
    6 &  & ST & Split-PoSh & 46.9 &78.3 & 87.5\\
    7 &  & ST & PGA & 44.1 & 74.2 & 84.5\\
    8 &  & ST & Progressive+PGA & 43.5 & 73.1 & 84.0 \\
    \bottomrule
  \end{tabular}
  }
  \caption{The ablation study of PSVT on 3DPW~\cite{von2018recovering} dataset. S: Spatial; ST: Spatio-Temporal. Conv, Split-PoSh, PGA and Progressive represent convolutional head, splitting pose and shape decoder in video Transformer baseline, pose-guided attention and progressive PGA in the decoder of PSVT, respectively.}
  \label{tab:aba}
\end{table}

\textbf{Component-wise Analysis in STSD}
The STSD of PSVT includes four modules: Token Aligning (TA), Window Pose Self-Attention (W-PSA), Window Shape Self-Attention (W-SSA), and Pose-Guided Attention (PGA). The ablation studies of these components are shown in Table~\ref{tab:pga}. The pose-guided attention with TA, W-PSA, and W-SSA achieves the best results. 

\begin{table}
  \centering
  \renewcommand\tabcolsep{8pt}
  \resizebox{\columnwidth}{!}{
  \begin{tabular}{@{}c|ccc|ccc@{}}
    \toprule
    \# & TA & W-PSA & W-SSA & PA-MPJPE & MPJPE & MPVE \\
    \midrule
    1 & \XSolidBrush & & & 46.2 & 76.6 & 85.1\\
    2 &  &\XSolidBrush & & 46.4 & 77.1 & 85.8\\
    3 & & & \XSolidBrush& 46.0 &76.1 & 85.0\\
    \midrule
    4 &  & & & 45.7 & 75.5 & 84.9 \\
    \bottomrule
  \end{tabular}
  }
  \caption{The ablation study of main components in STSD of PSVT on 3DPW~\cite{von2018recovering}. \XSolidBrush means the component is removed. 
  }
  \label{tab:pga}
\end{table}

\begin{figure}[t]
  \centering
%   \fbox{\rule{0pt}{2in} \rule{0.95\columnwidth}{0pt}}
  \includegraphics[width=\linewidth]{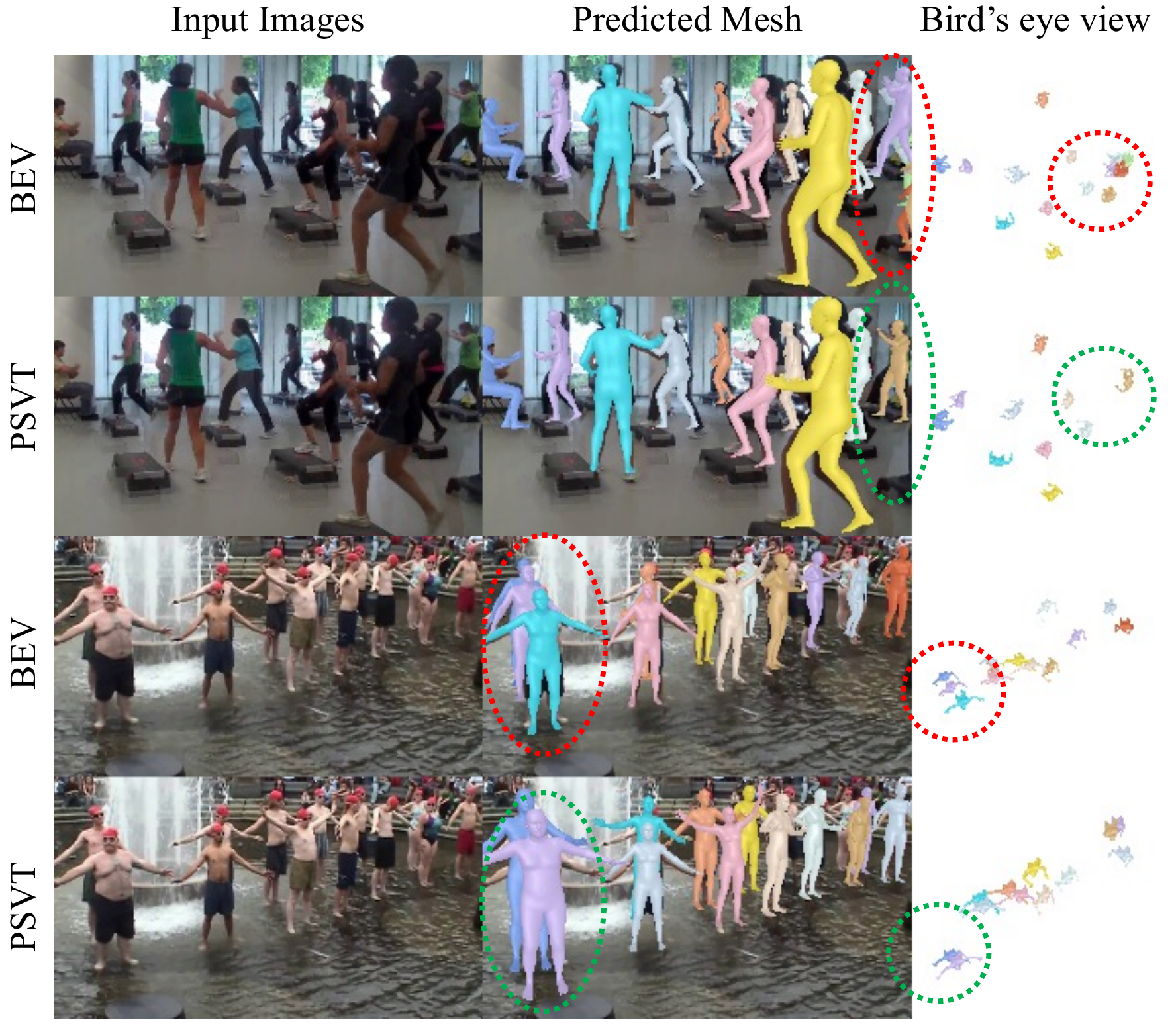}

  \caption{The comparison of BEV~\cite{sun2022putting} and PSVT. PSVT shows better predictions since PGA captures global interactions between spatial objects. Red circles indicate the wrong predictions.}
  \label{fig:mesh1}
\end{figure}

\begin{figure}[t]
  \centering
  \includegraphics[width=\linewidth]{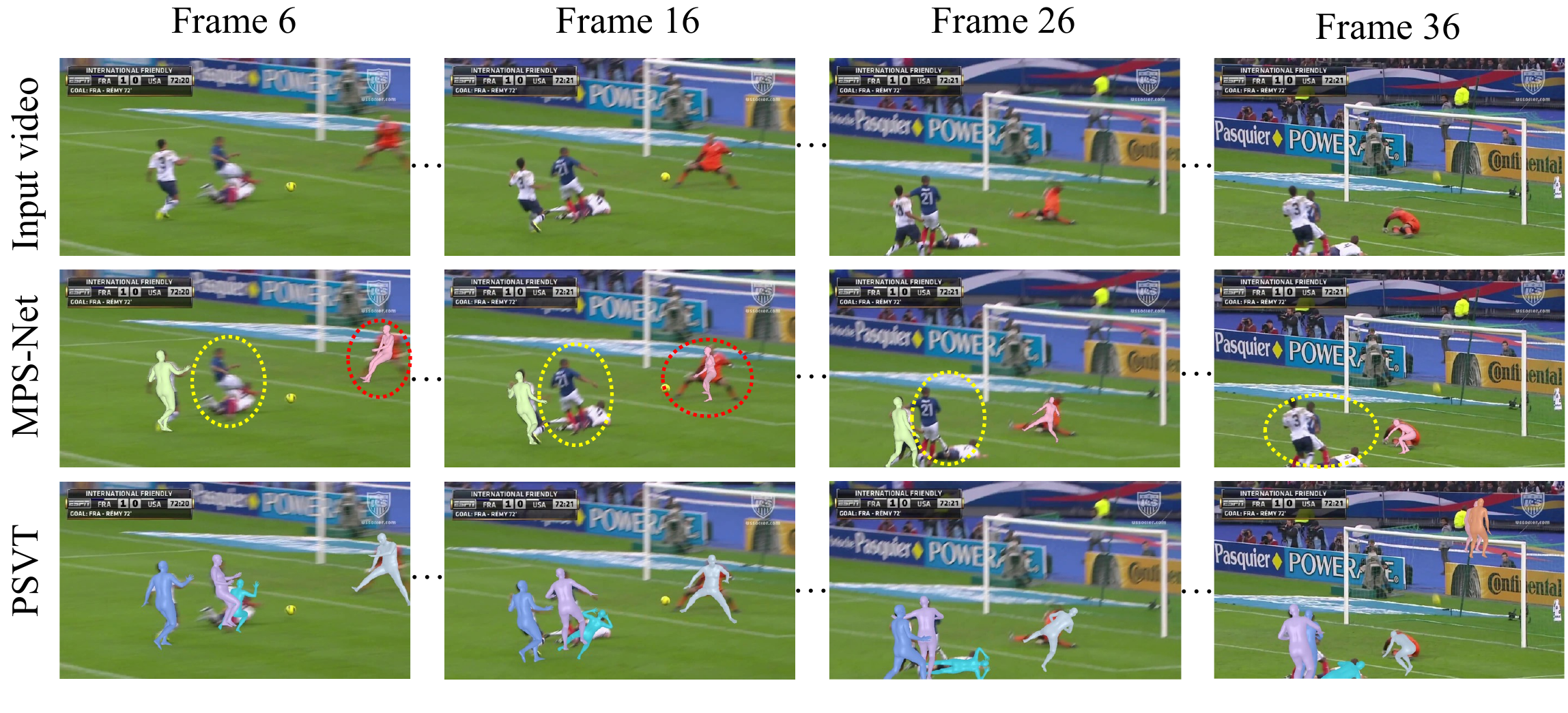}

  \caption{The comparison between MPS-Net~\cite{wei2022capturing} and our PSVT. PSVT shows better predictions since the captured global spatio-temporal context. Yellow circles and red circles indicate the missing objects and wrong predictions, respectively.}
  \label{fig:mesh2}
\end{figure}

\textbf{Visualization Analysis}
To evaluate the generalization ability of PSVT on the in-the-wild videos, we test PSVT on the videos from PoseTrack~\cite{andriluka2018posetrack} dataset. We compare PSVT with the image-based method and video-based method. 
Compared with BEV~\cite{sun2022putting}, as shown in Figure \ref{fig:mesh1}, PSVT shows better results in crowded scenarios due to the stronger ability to model global spatial context.
Compared with MPS-Net~\cite{wei2022capturing}, as shown in Figure \ref{fig:mesh2}, PSVT performs better results on the video since PSVT has stronger ability of modeling global spatio-temporal context. 
% More visualization results and analysis are shown in supplementary materials.

\section{Conclusion}
In this paper, we propose PSVT, the first end-to-end progressive video Transformer for multi-person 3D human pose and shape estimation, which captures the global context dependencies among different objects in both spatial and temporal dimensions. 
To handle the variances of objects as time proceeds and improve the performance of mesh estimation, we propose a progressive decoding mechanism and pose-guided attention for PSVT. Extensive experiments on four datasets show the effectiveness of proposed components, which leads PSVT to achieve state-of-the-art results.

%-------------------------------------------------------------------------
% \newpage

% \newpage

%%%%%%%%% REFERENCES
{\small
\bibliographystyle{ieee_fullname}
\bibliography{ref}
}

\begin{appendix}
\newpage
\section*{Supplementary Material}
In this supplementary material, we first give the algorithm details of PSVT in Section \ref{sec:alg}. Then, we will provide more experimental results and analysis of PSVT on the crowded scenarios in Section \ref{sec:crowd}. Then, some failure cases and the limitations of PSVT are discussed in Section \ref{sec:fail}. 
Finally, more visualization results on in-the-wild images or videos are shown in Section \ref{sec:vis}.

\section{Algorithm Details}
\label{sec:alg}

\begin{algorithm}[h]\small
  \caption{PSVT with progressive decoding mechanism and pose-guided attention.} % 名称
  \label{alg}
  \begin{algorithmic}[1]
    \Require
    % \REQUIRE 
      Video $\mathbf{V}$: $\{I^{t}, t \in [1,T]\}$;
      Backbone network of HRNet-32: $\phi(\cdot)$;
      Spatio-Temporal Encoder: $\text{STE}(\cdot)$;
      Spatio-Temporal Pose Decoder: $\text{STPD}(\cdot)$;
      Spatio-Temporal Shape Decoder: $\text{STSD}(\cdot)$;
      Token Aligning: $\text{TA}(\cdot)$;
      Joints weights of projecting mesh to 3D joints: $\mathcal{W}$.
    \Ensure
      Human meshes $\mathcal{M}=\{\mathcal{M}^t_i|t\in [1,T], i\in [1,N]\}$; 3D joints $J=\{J^t_i|t\in [1,T], i\in [1,N]\}$;\\
      Initializing $\mathcal{Q}_{pose}$,$\mathcal{Q}_{shape}$;
      \State $F=\{F^t|t\in [1,T]\}=\{\phi(I^t)| t\in[1,T]\}$;
      \State $\tau_e= \{\tau^t_e|t\in[1,T]\} = \{\text{STE}(F^t)|t\in[1,T]\}$;
      \For{$t = 1$; $t<=T$; $t++$}
          \State Updating pose queries $\hat{\mathcal{Q}}_{pose} = \psi(\mathcal{Q}^t_{pose}, \tau^{t-1}_{pose})$
          \State $\tau^t_{pose} = \text{STPD}(\hat{\mathcal{Q}}_{pose}, \tau^t_e)$;
          \State Updating shape queries $\hat{\mathcal{Q}}_{shape} =  \psi(\mathcal{Q}^t_{shape}, \tau^{t-1}_{shape})$
          \State Token aligning $\hat{\mathcal{Q}}_{shape} = \text{TA}(\hat{\mathcal{Q}}_{shape}, \tau^t_{pose})$;
          \State $\tau^t_{shape} = \text{STSD}(\hat{\mathcal{Q}}_{shape}, \tau^t_e)$;
          
          \State Regressing joints maps: $M_{2D}$, $M_o$, and $M_d$ from $\tau^t_{pose}$;
          \State Localizing Top-N center points $P = \{(x_i, y_i, d_i)|i\in [1,N]\}$ from joints maps;
          \State Regressing shape maps: $M_s$ from $\tau^t_{shape}$;
          
          \State Decoding mesh $\mathcal{M}^t= \{\mathcal{M}^t_i(\theta, \beta, \alpha)|i\in[1,N]\}$ with the center points of $P$;
          
          \State Projecting 3D joints $J^t=\{\mathcal{W}\mathcal{M}^t_i|i\in[1,N]\}$;
          
      \EndFor

  \end{algorithmic}
\end{algorithm}

The algorithms details of PSVT with progressive decoding mechanism and pose-guided attention are shown in Algorithm \ref{alg}.
The backbone network is HRNet-32~\cite{sun2019deep}. 
The progressive decoding mechanism is a bidirectional propagation scheme, which includes forward propagation
and backward propagation. For clarity, only forward propagation is shown in Algorithm \ref{alg}.

\begin{figure}[]
  \centering
%   \fbox{\rule{0pt}{2in} \rule{0.95\columnwidth}{0pt}}
  \includegraphics[width=\columnwidth]{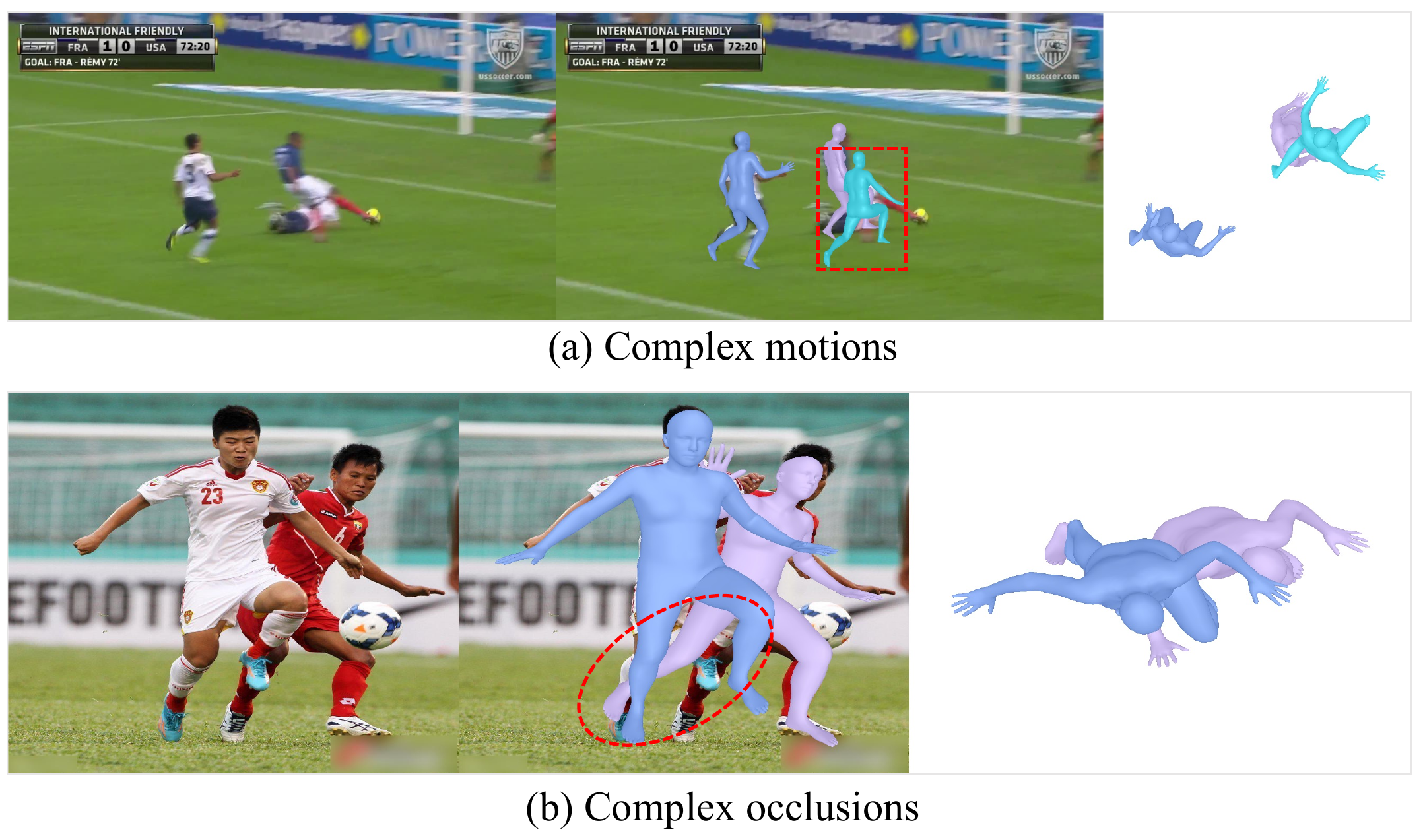}

  \caption{The visualization results of some failure cases on in-the-wild images. 
%   (a) Failure case of complex motions, (b) Failure cases of complex occlusions.
  }
  \label{fig:fail}
\end{figure}

\section{Evaluation on the Crowded Scenarios}
\label{sec:crowd}
To better evaluate the effectiveness of Pose-Guided Attention (PGA) in PSVT, we test PSVT on the crowded dataset. Following ~\cite{qiu2019learning,Zhang_2020_CVPR,sun2022putting,qiu2022ivt,qiu2023weakly}, we test PSVT on the occluded subset of 3DPW dataset~\cite{von2018recovering}. As shown in Table \ref{tab:oc}, we test PSVT without using temporal information for a fair comparison with BEV~\cite{sun2022putting}. PSVT achieves 49.67, 79.80, and 92.04 in PA-MPJPE, MPJPE, and MPVE, respectively. Compared with SOTA multi-person method (BEV), PSVT achieves relative gains of 7.2\%, 12.0\%, and 12.0\%, respectively. 
These results show that PSVT has a stronger ability to handle images with occluded persons since the PGA.

\begin{table}
  \centering
  \renewcommand\tabcolsep{16pt}
  \resizebox{\columnwidth}{!}{
  \begin{tabular}{c|c|ccc}
    \toprule
    Methods & Frame & PA-MPJPE & MPJPE & MPVE \\
    \midrule
    BEV~\cite{sun2022putting} & 1 & 53.55 & 90.64 & 104.55 \\
    \midrule
    \textbf{PSVT} (Ours) & 1& \textbf{49.67} & \textbf{79.80} & \textbf{92.04}\\
    \bottomrule
  \end{tabular}
  }
  \caption{The comparison between BEV~\cite{sun2022putting} and PSVT on the 3DPW-OC~\cite{Zhang_2020_CVPR}, the crowded subset of 3DPW~\cite{von2018recovering} dataset.
  }
  \label{tab:oc}
\end{table}

\begin{figure*}[]
  \centering

  \includegraphics[width=\linewidth]{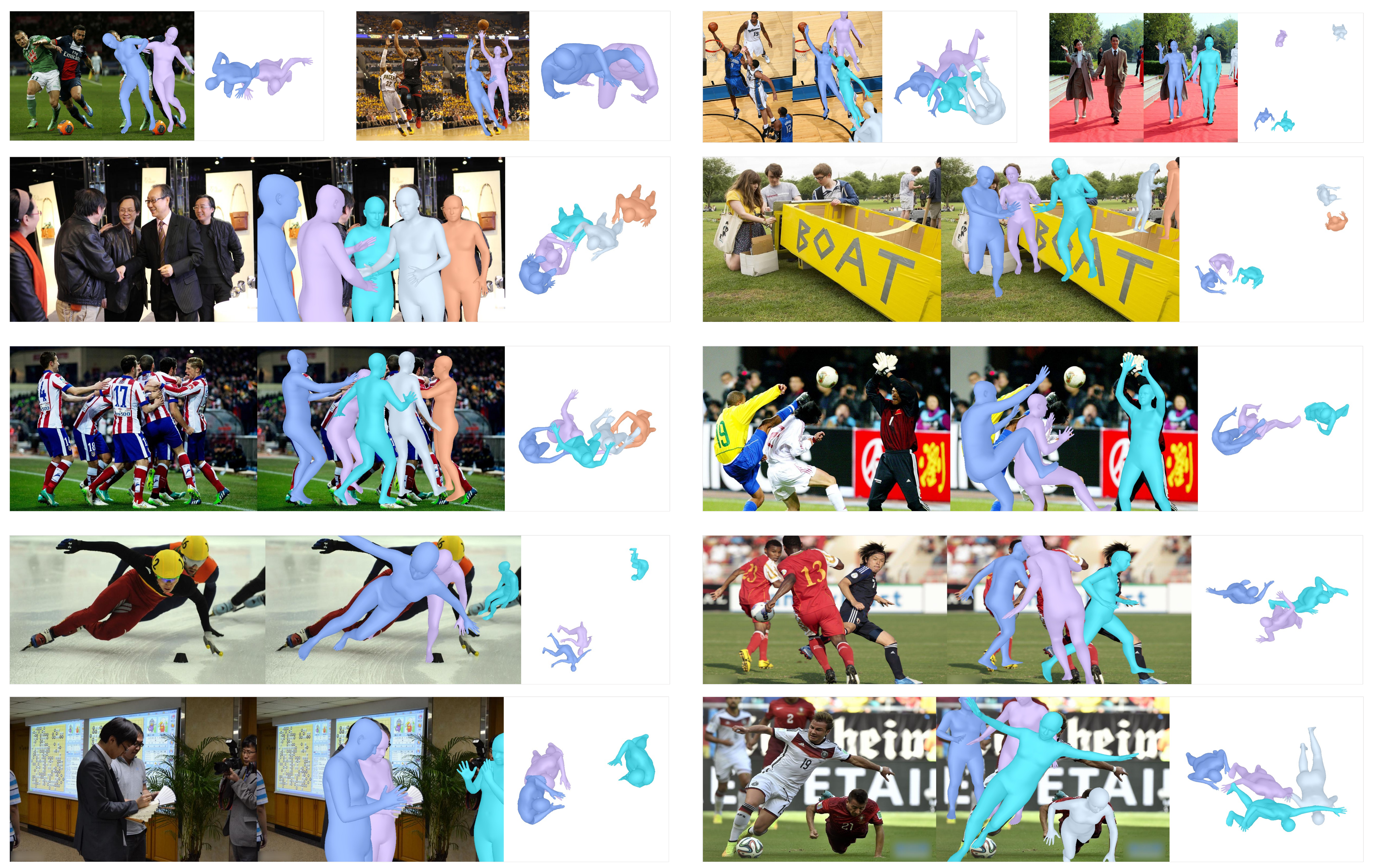}

  \caption{More visualization results on the in-the-wild images from CrowdPose~\cite{li2019crowdpose} dataset. More visualization results on in-the-wild videos are in the attached documents.}
  \label{fig:vis_img}
\end{figure*}

\section{Failure Cases and Limitations}
\label{sec:fail}
Although PSVT achieves state-of-the-art results on multiple widely-used 3D pose and shape estimation datasets, there still are some limitations and failure cases.
As shown in Figure ~\ref{fig:fail} (a), for some human instances with complex motions, it's difficult for PSVT to predict its pose and shape accurately.
As shown in Figure ~\ref{fig:fail} (b), for some human instances with complex occlusions, it's also difficult for PSVT to predict their poses and shapes accurately.

\section{More Visualization Results}
\label{sec:vis}

To evaluate the generalization ability of PSVT, we test PSVT on the in-the-wild images from CrowePose~\cite{li2019crowdpose} dataset and videos from PoseTrack~\cite{andriluka2018posetrack} dataset.
As shown in Figure \ref{fig:vis_img}, PSVT performs well on these images with crowded or strange poses, which shows the stronger generalization ability of PSVT.

%%%%%%%%% REFERENCES
% {\small
% \bibliographystyle{ieee_fullname}
% \bibliography{ref_s}
% }
\end{appendix}
\end{document}